\documentclass[10pt,twocolumn,letterpaper]{article}

\usepackage{ijcb}
\usepackage{times}
\usepackage{epsfig}
\usepackage{graphicx}
\usepackage{amsmath}
\usepackage{amssymb}
\usepackage{times}
\usepackage{epsfig}
\usepackage{graphicx}
\usepackage{amsmath}
\usepackage{amssymb}
\usepackage{booktabs}
\usepackage{multirow}
\usepackage{soul}
\usepackage{subfiles}
\usepackage[tableposition=top]{caption}
\usepackage[table,xcdraw]{xcolor}

\DeclareMathAlphabet{\pazocal}{OMS}{zplm}{m}{n}

\usepackage[pagebackref=true,breaklinks=true,letterpaper=true,colorlinks,bookmarks=false]{hyperref}

\ijcbfinalcopy 

\ifijcbfinal\fi
\begin{document}

\title{CoNAN: Conditional Neural Aggregation Network For Unconstrained Face Feature Fusion}

\author{Bhavin Jawade\thanks{Equal contribution authors in alphabetic order} \and Deen Dayal Mohan\footnotemark[1] \and Dennis Fedorishin \and Srirangaraj Setlur \and Venu Govindaraju \\
University at Buffalo, Buffalo, New York, USA \\
{\tt\small \{bhavinja,dmohan,dcfedori,setlur,govind\}@buffalo.edu}
}

\maketitle
\thispagestyle{empty}

\begin{abstract}

Face recognition from image sets acquired under unregulated and uncontrolled settings, such as at large distances, low resolutions, varying viewpoints, illumination, pose, and atmospheric conditions, is challenging. Face feature aggregation, which involves aggregating a set of N feature representations present in a template into a single global representation, plays a pivotal role in such recognition systems. Existing works in traditional face feature aggregation either utilize metadata or high-dimensional intermediate feature representations to estimate feature quality for aggregation. However, generating high-quality metadata or style information is not feasible for extremely low-resolution faces captured in long-range and high altitude settings.
To overcome these limitations, we propose a feature distribution conditioning approach called CoNAN for template aggregation. Specifically, our method aims to learn a context vector conditioned over the distribution information of the incoming feature set, which is utilized to weigh the features based on their estimated informativeness. The proposed method produces state-of-the-art results on long-range unconstrained face recognition datasets such as BTS, and DroneSURF, validating the advantages of such an aggregation strategy.
\end{abstract}

\vspace{-9px}

\section{Introduction}

\begin{figure}
\centering
\includegraphics[scale=0.155]{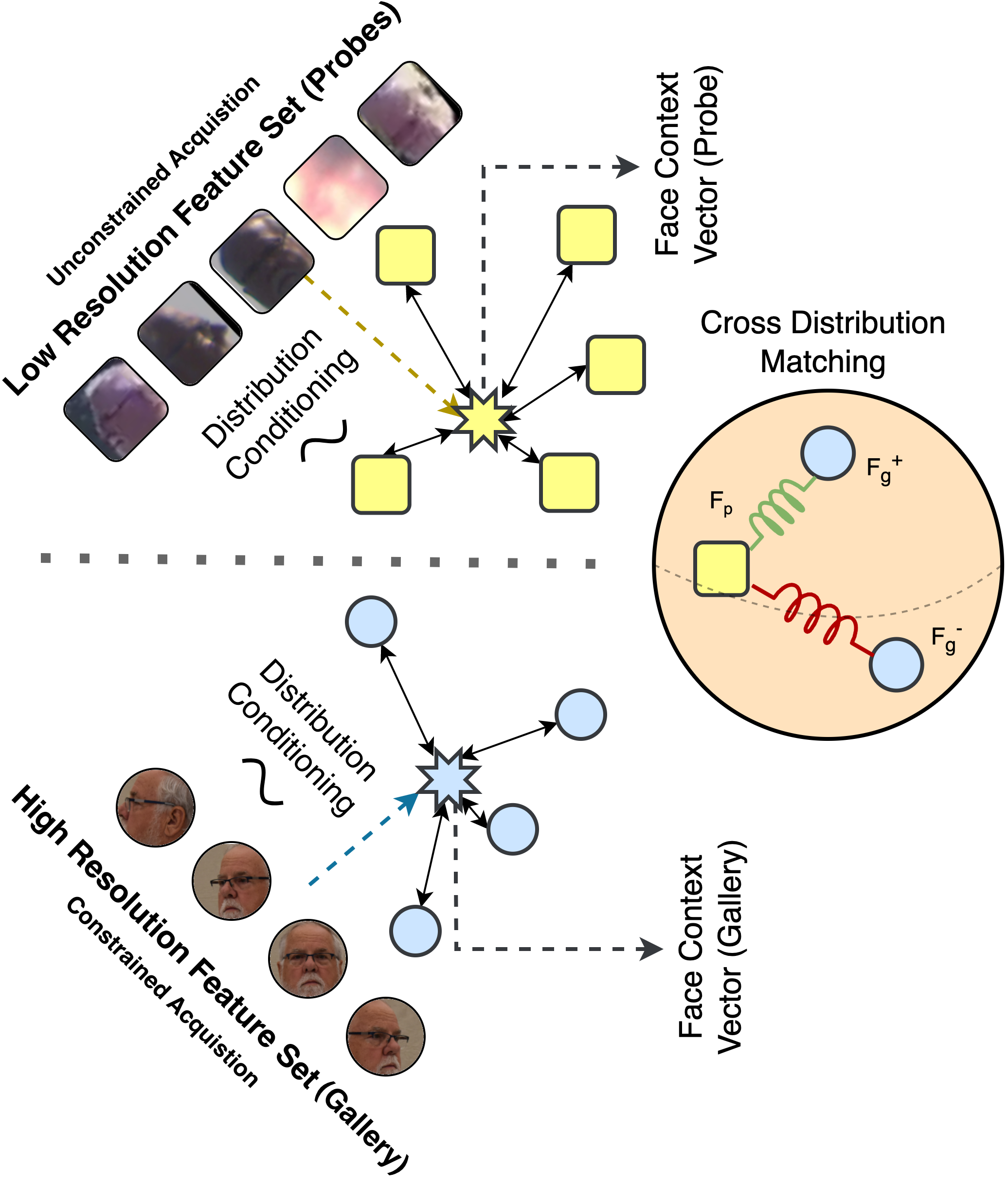}
\caption{Low resolution faces acquired from large distance and altitude have various kinds of distortions such as varying illumination, pose, atmospheric turbulence etc. Illustration provides an overview of how a distribution conditioned face context vector is utilized to compute informativeness of embeddings. (Best viewed in digital.}
\label{fig:conceptdiagram}
\vspace{-1.0em}
\end{figure}

Unconstrained face recognition has numerous applications in security, surveillance, biometric authentication, and social media analysis, among others. Tremendous progress has been made in developing systems that can automatically recognize subjects in the wild \cite{Arcface, Adaface}. Existing works in face recognition primarily focus on extracting low-dimensional discriminative feature representations from high-dimensional images. These methods include: i) Specialized \textit{margin-based loss functions} such as SphereFace \cite{Sphereface}, CosFace\cite{Cosface}, ArcFace \cite{Arcface}, and AdaCos \cite{Adacos} which utilize explicit margins to supplement the original softmax function,  ii) \textit{Mining-Based} and \textit{Curriculum-Based} loss functions such as MV-Arc-Softmax \cite{MV-Arc-Softmax}, CurricularFace \cite{Curricularface} and AdaFace \cite{Adaface} which explicitly emphasize the effects of hard and semi-hard samples and their order during training.

Although research in face recognition has primarily focused on creating highly discriminable face feature representation, another research direction is to develop methods for constructing optimal identity-specific templates from a set of face images. With the increasing availability of face data (e.g., from video surveillance and social media), face templates for a subject can now be created from multiple images and video frames. This has led to the natural evolution of face recognition to comparing sets of face images.

There has been increased interest recently in recognizing people at long-ranges and from high altitudes (\textit{e.g.} the IARPA BRIAR program). Unlike traditional face recognition where the subject is within a small distance from the camera, at larger distances and altitudes, the resolution of images tends to be very low. Subsequently, the inter-subject variance decreases, making identification challenging. Additionally, the intra-subject variance within a media (due to varying pose, distance, and atmospheric turbulence) increases. Often, the matching of images captured under this unconstrained setting has to be done with high quality gallery images captured under constrained settings. Figure \ref{fig:conceptdiagram} shows an example of typical samples in the probe and gallery sets for a subject. As one can note from the samples, the probe and gallery templates are of different data distributions and makes matching extremely challenging. 

Under these scenarios, face feature aggregation/fusion across multiple face images from different sources plays a pivotal role in achieving robust matching. Feature aggregation approaches
compute aggregation weights
for the samples in the sets depending on the importance of the image/feature. Although naive averaging can eliminate a fair amount of noise present in the samples, it does not weight features effectively based on their informativeness. Prior works in the literature  have proposed techniques such as Neural Aggregation Network (NAN) \cite{NAN}, which utilized cascaded attention blocks that analyze the feature vectors and assign weights for aggregation to produce a robust template representation. MCN \cite{MCN} used visual and content quality predicted from images for the creation of a unified representation. CAFace \cite{CAFace} proposed a two-stage feature aggregation method that also utilized high-dimensional intermediate feature-maps as style information during aggregation. TADPool \cite{TADPool} proposed  adapting a specific template's aggregation weights to the target template's properties using metadata information. 

Even though these methods have been successful, extracting high quality facial metadata (such as pose) or relevant style information from these low resolution and noisy face images captured under long-range and high altitude settings, is challenging.

Given this scenario, a good face feature aggregation technique must have the following properties: 
\begin{enumerate}
\itemsep0em 
\item It should adapt with varying number of features in the image-set
\item The method’s performance should not be conditioned on the availability of high-quality metadata or high dimensional intermediate feature maps from images
\item It should discount all the non-informative feature representations and prioritize highly discriminative feature embeddings
\item It should be able to adapt to a variety of face feature extractors with minimal retraining
\item The method should prioritize feature representation in the gallery that closely matches the distribution of probe features 
\item Should add minimal computational overhead to the existing feature representation

\end{enumerate}
Keeping these factors in mind, we design a novel approach called \textit{CoNAN - Conditional Neural Aggregation Network} to construct aggregated probe and gallery templates, conditioned on the template-level first-order statistics. Our intuition is that given a set of images in a probe template (captured under long range and high altitude setting), a majority of the media might not contain enough biometric information in order to contribute towards the final aggregated representation. Conditioning the aggregation strategy based on the first-order template level statistics will help provide a much more informative prior that could be leveraged by our aggregation network when computing attention weights. We conduct extensive experiments with the proposed aggregation method and provide both quantitative and qualitative results on BTS3.1 and DroneSURF (two long-range high-altitude unconstrained face recognition datasets).

\section{Related Works}


Face recognition is a well-explored area of research that has a significant real-world impact. The advent of deep learning techniques has significantly advanced the capabilities of face recognition systems in large-scale, long-range and unconstrained recognition scenarios. 

Designing novel loss formulations for face feature extraction are among the most explored avenues within face recognition. Some studies aim to design adaptive loss functions that utilize hard-mining \cite{MV-Arc-Softmax}, learning schedule based on sample difficultly \cite{Curricularface}, adaptive margin based on recognizability of the face \cite{MAGFace} or estimate sample importance based on image quality and margin \cite{Adaface}. Other works in the literature design margin based softmax loss functions \cite{Arcface, Sphereface, Cosface}. These works introduce different forms of cosine-margin functions such as \textit{angular} and \textit{additive} margins to provide better supervision to the feature extraction network.

An often necessary capability for a good face recognition system is the ability to effectively \textit{aggregate} multiple images to create a single representative feature embedding for an identity. Early approaches \cite{chen2015end, chowdhury2016one} relied on averaging and max-pooling to aggregate across multiple samples to create a unified representation. \cite{crosswhite2018template} improves upon these simple operations by learning an individual SVM for each identity to weight each sample that aggregates into a single representation. GhostVLAD \cite{zhong2019ghostvlad} is an end-to-end aggregation method that incorporates image characteristics such as quality, by mapping low-quality gallery images into a ``ghost" cluster. At the time of aggregation, the learned gallery image weights, if close to the ``ghost" cluster, are effectively ignored during aggregation. 

Some works on face feature aggregation utilize high-dimensional intermediate feature maps during aggregation because these feature maps provide rich and complementary information for feature reconstruction. CAFace \cite{CAFace} proposed a two-stage feature aggregation method that utilized high-dimensional intermediate feature-maps as style information during aggregation. The cluster stage is a linear assignment of $N$ inputs to $M$ global cluster centers, and the aggregation stage is a fusion over $M$ clustered features. RSA \cite{RSA} utilized residual self-attention to restructure the features using the other features within a set. To account for the computational complexity of inner and inter-set correlations, they proposed a permutation-invariant feature restructuring module.

Another line of work consists of methods that use metadata information to perform face fusion. MFAN \cite{MFAN} introduced a metadata-based feature aggregation method that utilized orthogonal data such as yaw, pitch, face size, etc. along with a siamese network to estimate strong correlations between the relative quality of the face images in a set. TADPool \cite{TADPool} further builds upon the use of metadata for feature aggregation by \textit{adapting} the probe and gallery together, based on characteristics such as face yaw and roll. An attention block is designed to pool probe features based on their compatibility to selected gallery features. \cite{NAN} proposed a network structure of two cascaded attention blocks to determine feature importance from an image set and used these importance scores to aggregate features. \cite{MCN} proposed a multicolumn network that weights an image within a set using its visual quality estimated using a self-quality assessment module, and then dynamically calibrates weights using its relative content quality with respect to other features images in the set. Our proposed method is similar to these works \cite{NAN, MCN} that aim to develop a foundational aggregation strategy that works directly on the low-dimensional embeddings without utilizing any metadata or intermediate map.

\section{Approach}

The overall objective of our method is similar to that of prior works in the area \cite{NAN, MCN}, where the network  has to automatically determine the optimal subset of face images in a template, given a target template to be compared with. We denote the two templates used for pooling as \textit{gallery} and \textit{probe} templates.

Let us consider a gallery set of face images $S_g = \{x_1, x_2, .., x_N\}$ where $x_1,..x_N$ are ($d$-dimensional) feature vectors representing the $N$ face images in the set.  Similarly, for a probe set of face images, we have $S_p = \{y_1, y_2, .., y_M\}$ feature vectors, for $M$ images in the set. We compute the pooled gallery template vector $T_g$ and the pooled probe template vector $T_p$ as

\begin{equation}
\label{eq:tg_conan}
T_g = \frac{\sum_{n=1}^N \vec{f}(\vec{g}(S_g))_n x_n}{\sum_{n=1}^N \vec{f}(\vec{g}(S_g))_n} 
\end{equation}

\begin{equation}
\label{eq:tp_conan}
T_p = \frac{\sum_{i=1}^M \vec{f}(\vec{g}(S_p))_n y_m}{\sum_{i=1}^M \vec{f}(\vec{g}(S_p))_n} 
\end{equation}

Here $T_g$ and $T_p$ are the normalized weighted average of $N$ gallery features ($x_n$) and $M$ probe features ($y_m$) respectively. $\vec{f} \in \mathbb{R}^{N}$ is a vector function that outputs an aggregation weight vector, and $\vec{f}(..)_n$ indexes into the $n^{th}$ term in the vector output. $\vec{g}$ is a function that computes first-order statistic vectors based on the features in $S_g$ and $S_p$.  Given feature sets $S_g$ and $S_p$ we compute the minimum, maximum, mean, variance, mode, and median along each dimension. 
\begin{equation}
\begin{split}
\vec{g}(S_g) = &\{max(S_g),min(S_g),mean(S_g), \\
&var(S_g),mode(S_g),median(S_g)\}
\end{split}
\end{equation}
Similarly for the probe template:
\begin{equation}
\begin{split}
\vec{g}(S_p) = &\{max(S_p),min(S_p),mean(S_p), \\
&var(S_p),mode(S_p),median(S_p)\}
\end{split}
\end{equation}
Computing $\vec{g}$ and conditioning $\vec{f}$ on $\Vec{g}$ has multiple advantages. First, the input to the aggregation function becomes fixed size. For example,  if we compute $K$ different first-order statistic vectors, then the input size to the aggregation function is $K$$\times$$d$ where $d$ is the dimension of each feature vector in the template. Secondly, the first-order statistics vectors help us capture a notion of the diversity of feature representations in the template, which is useful when creating an aggregated representation \cite{tulyakov2019utilizing}. Finally, the computation of $\vec{g}$ does not add significant computational overhead to the aggregation strategy.

\begin{figure*}[t]
\centering
\includegraphics[width=0.75\textwidth]{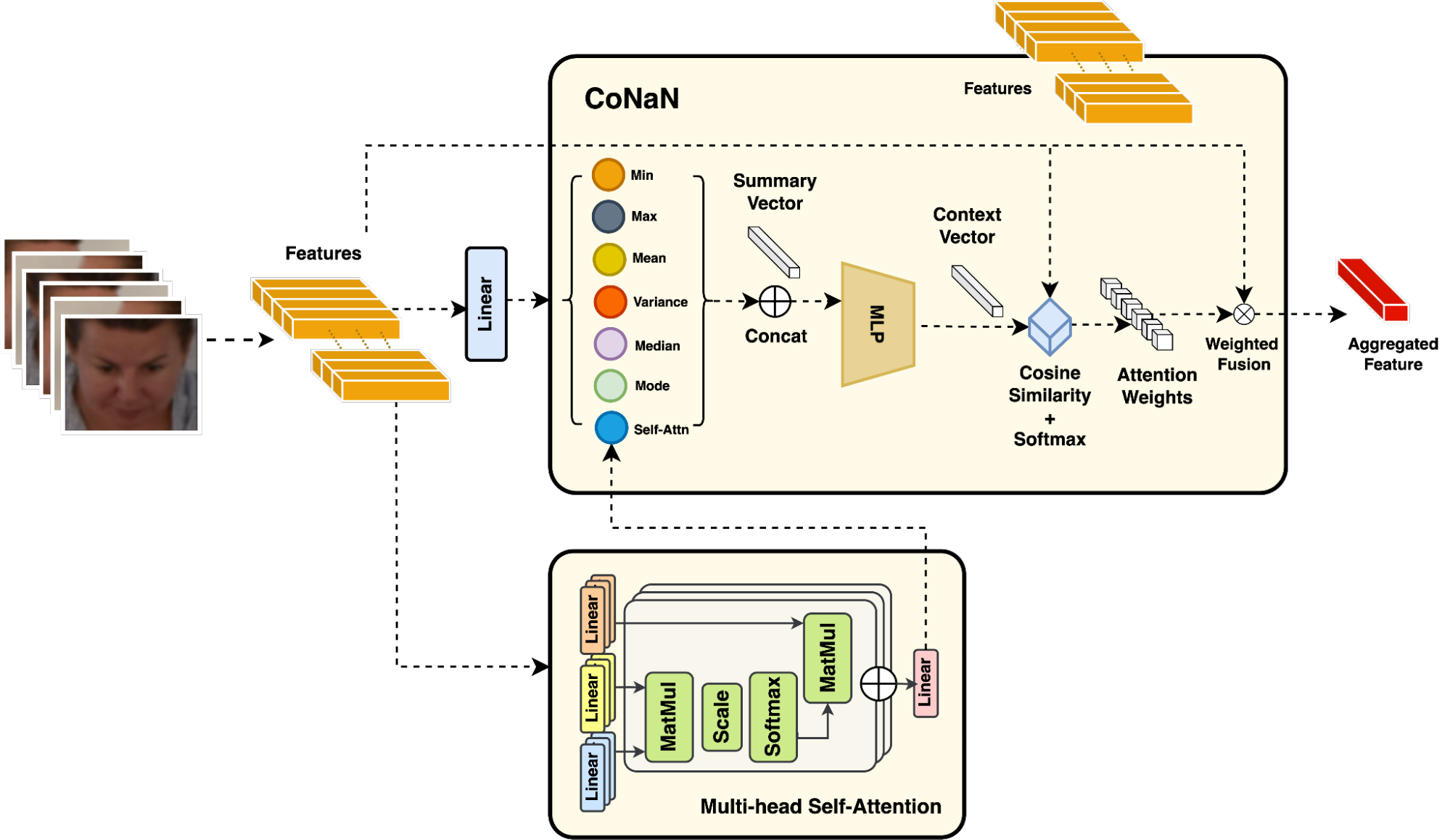}
\vspace{0.4em}
\caption[CoNAN Network Architecture ]{Training setup with the proposed CoNAN model. The setup is replicated as a siamese network, that aggregate both probe and gallery template.}
\label{fig:conan_arch}
\end{figure*}

One potential limitation of only using template-level statistics to condition the aggregation function is that it does not capture any other interactions between the feature representations in the template. These interactions can provide complementary information to the group-level information obtained by computing the first-order statistics. One way to incorporate these interactions is by using a scaled dot product attention operation.
Given the feature representations $\{x_1,......x_n\}$, we can project these feature representations using separate projection layers to create two tensors $K_x$ and $Q_x$. A matrix $\beta$ can be computed as an outer product of the tensors $K_x$ and $Q_x$ given as 
\begin{equation}
\label{attention}
\begin{split}
\beta = softmax(\frac{K_x \odot Q_x}{\sqrt{d}} )
\end{split}
\end{equation}

Given $\beta$, another projection layer, denoted as $V_x$, is used to capture the attended representation obtained by projecting $\beta$ with $V_x$.  

One limitation of directly applying the attention mechanism over the feature vectors in the template is that it does not account for variability in template size. Each template will have a varying number of face features, leading to a variable number of output vectors $V_x$. Since our aggregation method should be agnostic to the template size, we need to obtain a fixed-size output from the attention block. In order to obtain a fixed-size representation, we follow the approach of using a class token as described in \cite{dosovitskiy2020image}. A single learnable token is added as a new feature representation to every template. Only the representation corresponding to this token is selected from the set of vectors obtained after the projection of $\beta$ with $V_x$. This token representation is added to the set of template-level statistics. The updated $\vec{g}(S_g)$ is given by:

\begin{equation}
\label{ag_input_probe}
\begin{split}
\vec{g}(S_p) = &\{C_p, DTE_p, max(S_p),min(S_p),mean(S_p),\\
&var(S_p),mode(S_p),median(S_p)\}
\end{split}
\end{equation}
\begin{equation}
\label{ag_input_gallery}
\begin{split}
\vec{g}(S_g) =  &\{C_g, DTE_g, max(S_g),min(S_g),mean(S_g),\\
&var(S_g),mode(S_g),median(S_g)\}    
\end{split}
\end{equation}
$\text{DTE}_p$ and $\text{DTE}_g$ (Distribution Type Embedding) are learnable tokens that represent the distribution type of the current image set (probe or gallery). $C_g$ and $C_p$ represent these tokens after they are transformed using the attention module. We incorporate both $DTE_p$ (or $DTE_g$)  and $C_p$ (or $C_g$) to build $\vec{g}(S_p)$. We introduce separate learnable tokens for probe and gallery in the form of DTE so as to aid the aggregation function $\vec{f}$ and the multihead attention block to differentiate between probe and gallery templates and learn a distribution (gallery or probe) specific context vector.  

\subsection{Aggregation Network and Metric Learning}

The overall architecture of our approach is given in Figure \ref{fig:conan_arch}.

The aggregation function $\vec{f}$ is implemented as a fully connected network (FCN).  As shown in Figure \ref{fig:conan_arch},
the input to the network is the output of $\vec{g}$ as given by \ref{ag_input_gallery} and \ref{ag_input_probe}. 

The output of the aggregation network is a single $d$ dimensional vector called a context vector. Intuitively, one can think of a context vector as a representation that captures all the necessary information needed to  generate pooling weights for the template. Separate context vectors are  generated, for both probe and gallery templates, although the weights of the aggregation network are shared. 

Given the individual context vectors corresponding to the probe and gallery templates, the next step is to find weights that have to be assigned to each feature representation in the template.  In order to find these weights, we first find the cosine similarity of each feature representation in a template to the corresponding context vector. 

\begin{equation}
sim(V_g, S_g) = \frac{1}{N} \sum_{i=1}^{N} \frac{V_g \cdot x_i}{||V_g||\ ||x_i||}
\end{equation}

Here $V_g$ is the context vector corresponding to the gallery. This similarity value will indicate how informative each feature representation is, in the gallery template. A similar procedure is carried out to get  similarity values corresponding to the probe templates.  Now, given the set of similarity scores $sim(V_g, S_g)$, the attention weights corresponding to each feature representation are obtained by computing a weighted softmax on these similarity scores. This can be written as:

\begin{equation}
Wg_i = \frac{\exp(sim(V_g,S_g)_i / T)}{\sum_{i=1}^{N} \exp(sim(V_g, S_g)_i / T)}
\end{equation}

where $Wg_i$ is the weightage assigned to the $i^{th}$ feature representation in the template based on its similarity to the context vector. The temperature parameter $T$ controls the \textit{softness} of the softmax distribution, with higher values making the distribution more uniform and lower values concentrating the weightage more on the most similar vectors. We tune  temperature parameters so as to make the weight distribution sharper, thereby concentrating the weightage on a few of the most similar vectors. This follows our intuition that the majority of the feature representations in the template (especially probe) do not contain discriminative information, and hence in order to focus on the few representations with more discriminative power, we use a sharper softmax distribution. The final aggregated representation for gallery and probe templates is given by:

\begin{equation}
T_g = \sum_{i=1}^{N} Wg_i x_i
\end{equation}
\begin{equation}
T_p = \sum_{i=1}^{M} Wp_i y_i
\end{equation}

So far, we have investigated the network architecture of the aggregation strategy. We highlight that the weights of the fully connected layers are shared between the probe and gallery. This is done so  as to help the optimization process. 

Traditionally, aggregation methods are trained using a classification loss \cite{NAN} or MSE loss. In this work, we take a different approach by using metric learning methods. As mentioned in the previous sections, since the probe and gallery images are captured under extremely unconstrained settings, there is a high intra-class and low inter-class variance. Hence, an effective strategy would be to use a metric learning-based optimization. 

We use a loss formulation similar to \cite{khosla2020supervised} given by:

\begin{equation}
\mathcal{L} = -\frac{1}{|P(i)|} \sum_{p \in P(i)} \log \frac{\exp \left( \frac{z_i \cdot z_p}{\tau} \right)}{\sum_{a \in A(i)} \exp \left( \frac{z_i \cdot z_a}{\tau} \right)}
\end{equation}

Here $|P(i)|$ and $|A(i)|$ are the positive and negative samples in the current batch. If $z_i$ denotes an aggregated template representation ($T_p$ or $T_g$) in a batch, then $z_p$ refers to its mated template.  $z_a$ is an aggregated representation of a non-mated template. $\tau$ represents the temperature parameter in the metric learning formulation, which is used to find informative samples by adapting to different levels of noise and ambiguity in the data.   

\begin{table*}[h!]
\centering
\caption{Verification Performance (TAR (\%) @FAR=\%) for face included treatment and control protocols of the BTS 3.1 dataset.}
\vspace{0.8em}
\begin{tabular}{c|c|cccc||cccc}
\hline
\textbf{}      & \textbf{}         & \multicolumn{4}{c||}{\textbf{Face Included Treatment}}                                                                                                                                                                                                                        & \multicolumn{4}{c}{\textbf{Face Included Control}}                                                                                                                                                                                                                        \\ \hline
\rowcolor[HTML]{FFFFFF} 
\textbf{}      & Feature Extractor & \multicolumn{1}{c|}{\cellcolor[HTML]{FFFFFF}\textbf{$10^{-1}$}} & \multicolumn{1}{c|}{\cellcolor[HTML]{FFFFFF}\textbf{$10^{-2}$}} & \multicolumn{1}{c|}{\cellcolor[HTML]{FFFFFF}\textbf{$10^{-3}$}} & \textbf{$10^{-4}$} & \multicolumn{1}{c|}{\cellcolor[HTML]{FFFFFF}\textbf{$10^{-1}$}} & \multicolumn{1}{c|}{\cellcolor[HTML]{FFFFFF}\textbf{$10^{-2}$}} & \multicolumn{1}{c|}{\cellcolor[HTML]{FFFFFF}\textbf{$10^{-3}$}} & \textbf{$10^{-4}$} \\ \hline
\hline
\rowcolor[HTML]{FFFFFF} 
GAP  \cite{GAP}          & Arcface \cite{Arcface}           & \multicolumn{1}{c|}{\cellcolor[HTML]{FFFFFF}53.7}                            & \multicolumn{1}{c|}{\cellcolor[HTML]{FFFFFF}37.01}                           & \multicolumn{1}{c|}{\cellcolor[HTML]{FFFFFF}27.28}                           & 19.48                           & \multicolumn{1}{c|}{\cellcolor[HTML]{FFFFFF}91.17}                           & \multicolumn{1}{c|}{\cellcolor[HTML]{FFFFFF}84.82}                           & \multicolumn{1}{c|}{\cellcolor[HTML]{FFFFFF}75.26}                           & 66.21                           \\ \hline
\rowcolor[HTML]{FFFFFF} 
NAN \cite{NAN}     & Arcface \cite{Arcface}           & \multicolumn{1}{c|}{\cellcolor[HTML]{FFFFFF}55.41}                           & \multicolumn{1}{c|}{\cellcolor[HTML]{FFFFFF}39.01}                           & \multicolumn{1}{c|}{\cellcolor[HTML]{FFFFFF}26.64}                           & 18.3                            & \multicolumn{1}{c|}{\cellcolor[HTML]{FFFFFF}91.34}                           & \multicolumn{1}{c|}{\cellcolor[HTML]{FFFFFF}84.31}                           & \multicolumn{1}{c|}{\cellcolor[HTML]{FFFFFF}72.9}                            & 60.37                           \\ \hline
\rowcolor[HTML]{FFFFFF} 
MCN   \cite{MCN}         & Arcface \cite{Arcface}           & \multicolumn{1}{c|}{\cellcolor[HTML]{FFFFFF}55.06}                           & \multicolumn{1}{c|}{\cellcolor[HTML]{FFFFFF}39.41}                           & \multicolumn{1}{c|}{\cellcolor[HTML]{FFFFFF}28.22}                           & 19.37                           & \multicolumn{1}{c|}{\cellcolor[HTML]{FFFFFF}92.41}                           & \multicolumn{1}{c|}{\cellcolor[HTML]{FFFFFF}87.12}                           & \multicolumn{1}{c|}{\cellcolor[HTML]{FFFFFF}77.9}                            & 67.17                           \\ \hline
\rowcolor[HTML]{FFFFFF} 
\textbf{CoNAN} & Arcface \cite{Arcface}           & \multicolumn{1}{c|}{\cellcolor[HTML]{FFFFFF}\textbf{60.36}}                  & \multicolumn{1}{c|}{\cellcolor[HTML]{FFFFFF}\textbf{43.38}}                  & \multicolumn{1}{c|}{\cellcolor[HTML]{FFFFFF}\textbf{32.14}}                  & \textbf{23.14}                  & \multicolumn{1}{c|}{\cellcolor[HTML]{FFFFFF}\textbf{93.36}}                  & \multicolumn{1}{c|}{\cellcolor[HTML]{FFFFFF}\textbf{87.57}}                  & \multicolumn{1}{c|}{\cellcolor[HTML]{FFFFFF}\textbf{80.94}}                  & \textbf{71.89}                  \\ \hline
\hline

\rowcolor[HTML]{FFFFFF} 
GAP   \cite{GAP}         & Adaface \cite{Adaface}           & \multicolumn{1}{c|}{\cellcolor[HTML]{FFFFFF}63.79}                           & \multicolumn{1}{c|}{\cellcolor[HTML]{FFFFFF}50.76}                           & \multicolumn{1}{c|}{\cellcolor[HTML]{FFFFFF}40.81}                           & 31.7                            & \multicolumn{1}{c|}{\cellcolor[HTML]{FFFFFF}96.17}                           & \multicolumn{1}{c|}{\cellcolor[HTML]{FFFFFF}91.28}                           & \multicolumn{1}{c|}{\cellcolor[HTML]{FFFFFF}86.9}                            & 80.1                            \\ \hline

\rowcolor[HTML]{FFFFFF} 
CAFace*  \cite{CAFace}         & Adaface \cite{Adaface}           & \multicolumn{1}{c|}{\cellcolor[HTML]{FFFFFF}61.82}                           & \multicolumn{1}{c|}{\cellcolor[HTML]{FFFFFF}51.31}                           & \multicolumn{1}{c|}{\cellcolor[HTML]{FFFFFF}41.95}                           & 33.41                           & \multicolumn{1}{c|}{\cellcolor[HTML]{FFFFFF}95.55}                           & \multicolumn{1}{c|}{\cellcolor[HTML]{FFFFFF}92.86}                           & \multicolumn{1}{c|}{\cellcolor[HTML]{FFFFFF}88.75}                           & 82.96                           \\ \hline

\rowcolor[HTML]{FFFFFF} 
NAN    \cite{NAN}        & Adaface \cite{Adaface}           & \multicolumn{1}{c|}{\cellcolor[HTML]{FFFFFF}65.29}                           & \multicolumn{1}{c|}{\cellcolor[HTML]{FFFFFF}54.44}                           & \multicolumn{1}{c|}{\cellcolor[HTML]{FFFFFF}44.96}                           & 34.86                           & \multicolumn{1}{c|}{\cellcolor[HTML]{FFFFFF}96.06}                           & \multicolumn{1}{c|}{\cellcolor[HTML]{FFFFFF}93.31}                           & \multicolumn{1}{c|}{\cellcolor[HTML]{FFFFFF}90.16}                           & 84.82                           \\ \hline
\rowcolor[HTML]{FFFFFF} 
MCN   \cite{MCN}         & Adaface \cite{Adaface}           & \multicolumn{1}{c|}{\cellcolor[HTML]{FFFFFF}65.22}                           & \multicolumn{1}{c|}{\cellcolor[HTML]{FFFFFF}54.25}                           & \multicolumn{1}{c|}{\cellcolor[HTML]{FFFFFF}45.01}                           & 34.84                           & \multicolumn{1}{c|}{\cellcolor[HTML]{FFFFFF}96.06}                           & \multicolumn{1}{c|}{\cellcolor[HTML]{FFFFFF}93.19}                           & \multicolumn{1}{c|}{\cellcolor[HTML]{FFFFFF}89.82}                           & 85.32                           \\ \hline
\rowcolor[HTML]{FFFFFF} 
\textbf{CoNAN} & Adaface \cite{Adaface}           & \multicolumn{1}{c|}{\cellcolor[HTML]{FFFFFF}\textbf{67.56}}                  & \multicolumn{1}{c|}{\cellcolor[HTML]{FFFFFF}\textbf{56.32}}                  & \multicolumn{1}{c|}{\cellcolor[HTML]{FFFFFF}\textbf{46.14}}                  & \textbf{36.52}                  & \multicolumn{1}{c|}{\cellcolor[HTML]{FFFFFF}\textbf{96.06}}                  & \multicolumn{1}{c|}{\cellcolor[HTML]{FFFFFF}\textbf{93.7}}                   & \multicolumn{1}{c|}{\cellcolor[HTML]{FFFFFF}\textbf{90.27}}                  & \textbf{85.72}                  \\ \hline
\end{tabular}
\label{tab:briar}
\vspace{0.2em}
\\ $\ast$ We benchmark CAFace model pretrained on WebFace4M released by the authors
\end{table*}
\begin{table}[h!]
\centering
\caption{Rank-1 accuracy (\%) for video-wise identification on DroneSURF dataset.}
\vspace{0.8em}
\begin{tabular}{cccc}
\hline
\multicolumn{4}{c}{\textbf{Trained On DroneSURF}}                                                                                   \\ \hline \hline

\multicolumn{1}{c|}{}               & \multicolumn{1}{c|}{Feature Extractor} & \multicolumn{1}{c|}{Active}         & Passive        \\ \hline \hline

\multicolumn{1}{c|}{HOG \cite{HOG}}            & \multicolumn{1}{c|}{-}                 & \multicolumn{1}{c|}{8.33}           & 7.30           \\ \hline
\multicolumn{1}{c|}{LBP \cite{LBP}}            & \multicolumn{1}{c|}{-}                 & \multicolumn{1}{c|}{4.16}           & 4.16           \\ \hline
\multicolumn{1}{c|}{VGGFace \cite{VGGFace}}        & \multicolumn{1}{c|}{-}                 & \multicolumn{1}{c|}{16.67}          & 5.20           \\ \hline
\multicolumn{1}{c|}{COTS \cite{droneSURF}}           & \multicolumn{1}{c|}{-}                 & \multicolumn{1}{c|}{21.88}          & 4.16           \\ \hline
\multicolumn{1}{c|}{GAP \cite{GAP}}            & \multicolumn{1}{c|}{Arcface \cite{Arcface}}           & \multicolumn{1}{c|}{16.67}          & 8.33           \\ \hline
\multicolumn{1}{c|}{\textbf{CoNAN \cite{NAN}}} & \multicolumn{1}{c|}{\textbf{Arcface \cite{Arcface}}}  & \multicolumn{1}{c|}{\textbf{17.71}} & \textbf{13.54} \\ 
\hline \hline
\multicolumn{1}{c|}{GAP \cite{GAP}}            & \multicolumn{1}{c|}{Adaface \cite{Adaface}}           & \multicolumn{1}{c|}{46.87}          & 7.29           \\ \hline
\multicolumn{1}{c|}{NAN \cite{NAN}}            & \multicolumn{1}{c|}{Adaface \cite{Adaface}}           & \multicolumn{1}{c|}{65.62}          & 6.25           \\ \hline
\multicolumn{1}{c|}{MCN \cite{MCN}}            & \multicolumn{1}{c|}{Adaface \cite{Adaface}}           & \multicolumn{1}{c|}{72.92}          & 8.33           \\ \hline
\multicolumn{1}{c|}{\textbf{CoNAN \cite{NAN}}} & \multicolumn{1}{c|}{\textbf{Adaface} \cite{Adaface}}           & \multicolumn{1}{c|}{\textbf{80.21}} & \textbf{13.54} \\ 
\hline
\hline
\multicolumn{4}{c}{\textbf{Trained On BRS (Cross-Dataset)}}                                                                                  \\ \hline
\multicolumn{1}{c|}{GAP \cite{GAP}}            & \multicolumn{1}{c|}{Adaface \cite{Adaface}}           & \multicolumn{1}{c|}{46.87}          & 7.29           \\ \hline
\multicolumn{1}{c|}{NAN \cite{NAN}}            & \multicolumn{1}{c|}{Adaface \cite{Adaface}}           & \multicolumn{1}{c|}{80.21}          & 8.33           \\ \hline
\multicolumn{1}{c|}{MCN \cite{MCN}}            & \multicolumn{1}{c|}{Adaface \cite{Adaface}}           & \multicolumn{1}{c|}{79.16}          & 10.41          \\ \hline
\multicolumn{1}{c|}{\textbf{CoNAN \cite{NAN}}} & \multicolumn{1}{c|}{\textbf{Adaface \cite{Adaface}}}  & \multicolumn{1}{c|}{\textbf{83.33}} & \textbf{12.50} \\ \hline
\end{tabular}
\label{tab:dronesurf}
\end{table}

\section{Experiments}
In this section, we will describe the datasets used, experimental setting and results obtained for the proposed method.

\subsection{Datasets}
To demonstrate the robustness of our aggregation method in the unconstrained matching scenario, we select challenging datasets with  low quality images and videos collected in the long range setting using a variety of sensors (such as UAVs and long range cameras). We use the following datasets for either training or evaluation in our experiments. 
\\
1. \textbf{BRIAR Research Set 3 (BRS 3)} \cite{briar_dataset} - This dataset was collected and released as a part of IARPA's Biometric Recognition and Identification at Altitude and Range (BRIAR) program Phase 1. BRS 3 comprises of videos and images that were gathered from 170 participants in both controlled and field scenarios. The controlled setting consists of high-resolution faces captured in a closed environment at close range, while the field media is captured in an unconstrained setting from both close range and long range, with distances varying from 100 meters to 500 meters. We utilized 49,429 video clips and images from BRS 3 to train CoNAN. This comprises of 20,780 video clips from the field setting and 23,489 video clips and 5,160 images from the controlled setting. In our training and validation procedures, we considered all controlled videos and images as the gallery, and all field videos as probes.
\\
2. \textbf{BRIAR Test Set 3.1 (BTS 3.1)} \cite{briar_dataset} - We evaluate our method on BTS 3.1 which was the test set for the evaluation protocols for IARPA BRIAR Phase 1. In this paper, we present results on the face included treatment and face included control protocols of the BTS 3.1 test set. \textit{Face included treatment} comprises of 5,822 probe videos from 260 subjects acquired in long-range uncontrolled and unconstrained setting while \textit{Face included control} comprises of 1,914 probe videos from 256 subjects acquired in a regulated long-range setting. Gallery associated with BTS 3.1 is divided into two parts, gallery 1 and gallery 2. Gallery 1 consists of 47,925 video clips and images from 485 subjects, and Gallery 2 consists of 47,413 video clips and images from 481 subjects. Both gallery 1 and gallery 2 contain a common set of 351 distractor identities.
\\ 
3. \textbf{DroneSURF Dataset} \cite{droneSURF} - The dataset contains 200 videos of 58 subjects, captured across 411K frames using drones. The dataset has been collected under two survelliance settings: (i) Active Surveillance and (ii) Passive Surveillance. Each setting consists of around 100 videos. Following \cite{droneSURF}, we divide the subjects into a random 40-60 split. 40\% of the subjects (24 identities) are kept for testing and 60\% (34 identities) are used for training and validation. The dataset is released with over 786K face annotations. To demonstrate the performance of our aggregation method, we report results only on the video-wise identification protocol of DroneSURF. 

\subsection{Implementation Details}
In this section, we will discuss the configuration of all the deep neural networks used in the proposed framework.
1. \textbf{Face Recognizer:} We follow previous works and use
MTCNN \cite{MTCNN} for face detection and alignment. Face representations are computed  using pre-trained Adaface \cite{Adaface} and Arcface \cite{Arcface} feature extractors. For Adaface \cite{Adaface} we use a ResNet-101 model pre-trained on WebFace 12M \cite{webface}. Each face image is resized to 112x112x3 before feature extraction. For Arcface \cite{Arcface}, we extract features using a MS1MV2 pre-trained ResNet-50 backbone.
\\
2. \textbf{CoNAN Architecture:} The MLP involved in the CoNAN architecture consists of 3 layers that non-linearly transform the summary vector of dimension 4096 ($512\times8$)
to a $1\times512$ dimensional context vector. The multi-head attention layer involved uses 4 heads. For both probes and gallery we use a softmax temperature of 0.067 for BRS 3 training and 0.1 for DroneSURF training. During training, we use a temperature of 0.1 for the supervised contrastive loss with a cross-batch memory. We validate and early-stop on 30 and 10 identities set aside from the training identities split for BRS 3.0 and DroneSURF respectively. For optimization, we utilize an Adam optimizer with learning rates for the MLP, CLS tokens and multi-head attention layer set to $1e^{-2}$, and for the probe linear transformation layer to $1e^{-4}$. 

\begin{figure*}
\centering
\includegraphics[scale=0.5]{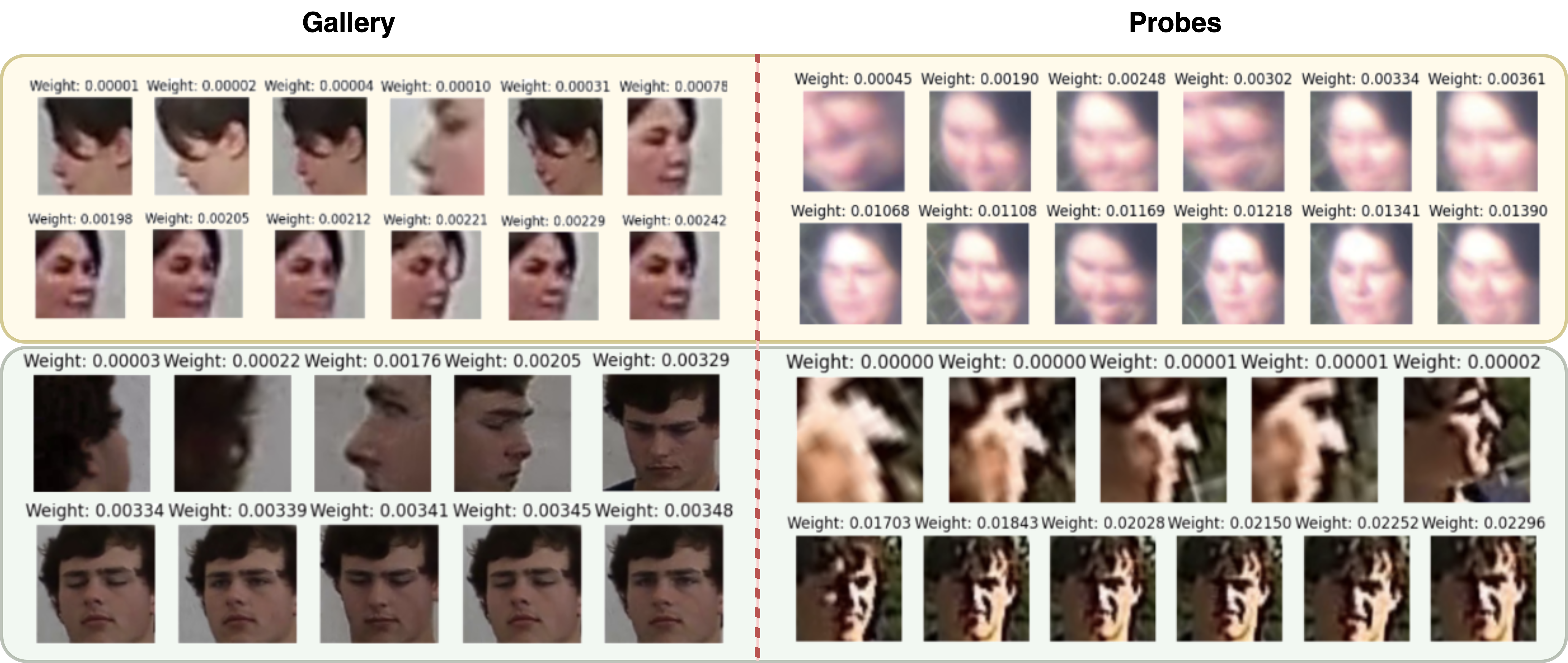}
\vspace{0.1em}
\caption{Qualitative results on BTS3.1 dataset's gallery and probe set. Images on the left are from high quality gallery, and images on the right are from low resolution long-range probes. Faces are sorted based on CoNAN attention weights from low to high.}
\label{fig:qualititave_diagram}
\vspace{-0.5em}
\end{figure*}

\subsection{Comparison to State of the Art}

We compare our fusion method against multiple face aggregation and feature extraction methods proposed in the literature. Table \ref{tab:briar} presents experiments on the BTS 3.1 \cite{briar_dataset} dataset with methods trained on the BRS 3 dataset \cite{briar_dataset}. In comparison to GAP (Global Average Pooling) \cite{GAP} or naive averaging, CoNAN performs nearly 6.47\% and 5.56\% better at $10^{-2}$ FAR with Arcface and Adaface embeddings respectively in the Face Included Treatment setting. In the Face Included Control setting, we observe a 2.75\% and 2.5\% improvement over GAP using Arcface and Adaface embeddings respectively at $10^{-2}$ FAR. It can be observed that CoNAN provides a substantial jump especially in the face included treatment protocol which is a more challenging scenario where faces are acquired in a fully unconstrained setting at much larger distances. In comparison to MCN (Multi-Column Networks) \cite{MCN}, CoNAN improves by 2.07\% at $10^{-2}$ FAR in the Face Included Treatment setting. It can also be observed that our aggregation method consistently improves across all FAR thresholds.

Table \ref{tab:dronesurf} shows results on the DroneSURF dataset. In the active survelliance setting using Arcface embeddings, CoNAN improves 1.05\% on GAP. With Adaface embeddings, we observe a 33.34\% improvement over GAP. With respect to MCN \cite{MCN}, there is a 7.29\% improvement using CoNAN. In passive survelliance setting, CoNAN improves 5.21\% over MCN. We also perform cross-dataset evaluations to demonstrate the robustness of CoNAN to distribution shifts. When trained on BRS 3 and evaluated on DroneSURF, CoNAN improves by 4.17\% and 2.09\% with respect to MCN (trained on BRS 3 and evaluted on DroneSURF) in active and passive setting respectively.

\subsection{Ablation and Discussion}
In this section, we discuss different properties of CoNAN using multiple qualitative and quantitative experiments:
\\
\textbf{Qualitative Analysis:} Figure \ref{fig:qualititave_diagram} shows a qualitative analysis on BTS 3.1\footnote{Subjects have consented to publish their BRIAR images.} of the feature importance score computed using CoNAN architecture. As can be observed in the gallery images of both identity 1 (top) and identity 2 (bottom), CoNAN successfully down-weights face images with minimal discriminative information, for example the face covered with hair, back of the head, side close-up images, etc. Similarly, in the probe images (on the right) it can be noticed that even though all face images are low resolution acquired from long-range, CoNAN down-weights faces that are relatively more blurry or have lesser discriminative facial features. It is worth noting that CoNAN is able to estimate the informativeness of these features without using any explicit metadata or high-dimensional intermediate features. An additional advantage of CoNAN's ability to discard bad features such as those extracted from inaccurate face detections, is that it allows for a lower threshold for face detectors thereby reducing FTE (Failure to enroll) cases.
\begin{table*}[]
\centering
\caption{An Ablation study on different components of the summary vector.}
\vspace{0.5em}
\begin{tabular}{l|l|l|l|l|l|l|l|l|l|l|l}
\hline
\textbf{Mean}             & \cellcolor[HTML]{FFFFFF}\textbf{Var} & \textbf{Max}              & \textbf{Min}              & \textbf{Mode}             & \textbf{Median}           & \textbf{MHA}              & \textbf{DTE}              & \multicolumn{1}{c|}{\textbf{$10^{-1}$}} & \multicolumn{1}{c|}
{\textbf{$10^{-2}$}} & \multicolumn{1}{c|}
{\textbf{$10^{-3}$}} & \multicolumn{1}{c}
{\textbf{$10^{-4}$}} \\ 
\hline
\hline
\checkmark &                                      &                           &                           &                           &                           &                           &                           &                                                                                                                                                                                       
65.82 & 55.05 &  45.66 &  35.86

\\ \hline
\checkmark & \checkmark            &                           &                           &                           &                           &                           &                           &     

65.95 & 55.24 & 45.52 & 35.63
\\ \hline
\checkmark & \checkmark            & \checkmark &                           &                           &                           &                           &                           &                                                      
66.29 & 55.64 & 45.70 & 35.88
\\ \hline
\checkmark & \checkmark            & \checkmark & \checkmark &                           &                           &                           &                           &                                                                
66.60 & 55.83 & 45.61 & 35.93 
\\ \hline
\checkmark & \checkmark            & \checkmark & \checkmark & \checkmark &                           &                           &                           &   
66.73 & 55.91 & 45.76 & 36.05
\\ \hline
\checkmark & \checkmark            & \checkmark & \checkmark & \checkmark & \checkmark &                           &                           &          66.56 & 56.02 & 45.84 & 36.05                                           \\ \hline
\checkmark & \checkmark            & \checkmark & \checkmark & \checkmark & \checkmark & \checkmark &                           &                                                      67.33 & 56.27 & 46.02 & 36.31                          \\ \hline
\checkmark & \checkmark            & \checkmark & \checkmark & \checkmark & \checkmark & \checkmark & \checkmark &  67.56 & 56.32 & 46.14 & 36.52                                                     \\ \hline
\end{tabular}
\label{tab:ablation}
\end{table*}
\\
\textbf{Cross Dataset Performance:} In Table \ref{tab:dronesurf}, we also present cross-dataset performance of CoNAN (Last three rows). In this experiment, CoNAN was trained using Adaface features on BRS dataset and evaluated on DroneSURF dataset. As can be observed, in both active and passive surveillance settings, CoNAN performs significantly better than MCN \cite{MCN}, NAN \cite{NAN} and GAP \cite{GAP}.    
\\
\textbf{Changing Feature Extractors:} In Table \ref{tab:briar} and \ref{tab:dronesurf}, we present results using both Arcface \cite{Arcface} and Adaface \cite{Adaface} feature extractors to demonstrate that CoNAN improves over the baseline performance irrespective of the feature extractor that is used and is agnostic to the feature extractor.
\\
\textbf{Summary Vector Component Ablation:} We present an ablation study on the different components of the summary vector in Table \ref{tab:ablation}. When conditioning using only mean information we achieve a 0.6\% and 0.8\% improvement at $10^-2$ FAR over \cite{NAN} and \cite{MCN}. Introducing the max and min information provides another 0.6\% jump in the performance over just using mean and variance. Finally, adding the multihead attention vector along with distribution type embedding (DTE) helps achieve the best results. It can also be observed here that multihead attention (MHA) alone is not sufficient to achieve best performance for face feature aggregation, but rather conditioning over all of the distribution information is necessary.

\section{Conclusion}
Face feature aggregation plays a pivotal role in unconstrained long-range face recognition. In this work, we explore a novel feature aggregation technique that does not rely on the availability of high dimensional intermediate feature or any form of metadata information to estimate informativeness of features. Specifically, we introduce the notion of conditioning over the distribution information to learn a context vector that can be utilized to compute feature weights. Through quantitative experiments on multiple datasets using multiple feature extractors we demonstrate that CoNAN improves significantly over existing approaches. We also present qualitative results to show that CoNAN is able to emphasize high quality features and discount non-discriminative facial features. Although we present CoNAN as a face feature aggreagation method, given that the architecture itself is independent of any specific modality, future works in this domain can explore applying CoNAN to aggregate features in other problems such as video-based person re-identification \cite{videobasedreid}, set-based fingerprint recognition \cite{Ridegbase, multilossfusion}, and other tasks. 

\section{Acknowledgment}
This research is based upon work supported in part by the Office of the Director of National Intelligence (ODNI), Intelligence Advanced Research Projects Activity (IARPA), via \textit{2022-21102100001}. The views and conclusions contained herein are those of the authors and should not be interpreted as necessarily representing the official policies, either expressed or implied, of ODNI, IARPA, or the U.S. Government. The U.S. Government is authorized to reproduce and distribute reprints for governmental purposes notwithstanding any copyright annotation therein.

{\small
\bibliographystyle{ieee_fullname}
\bibliography{egbib}

\begin{thebibliography}{10}\itemsep=-1pt

\bibitem{chen2015end}
Jun-Cheng Chen, Rajeev Ranjan, Amit Kumar, Ching-Hui Chen, Vishal~M Patel, and
  Rama Chellappa.
\newblock An end-to-end system for unconstrained face verification with deep
  convolutional neural networks.
\newblock In {\em Proceedings of the IEEE international conference on computer
  vision workshops}, pages 118--126, 2015.

\bibitem{chowdhury2016one}
Aruni~Roy Chowdhury, Tsung-Yu Lin, Subhransu Maji, and Erik Learned-Miller.
\newblock One-to-many face recognition with bilinear cnns.
\newblock In {\em 2016 IEEE Winter Conference on Applications of Computer
  Vision (WACV)}, pages 1--9. IEEE, 2016.

\bibitem{briar_dataset}
David Cornett, Joel Brogan, Nell Barber, Deniz Aykac, Seth Baird, Nicholas
  Burchfield, Carl Dukes, Andrew Duncan, Regina Ferrell, Jim Goddard, et~al.
\newblock Expanding accurate person recognition to new altitudes and ranges:
  The briar dataset.
\newblock In {\em Proceedings of the IEEE/CVF Winter Conference on Applications
  of Computer Vision}, pages 593--602, 2023.

\bibitem{crosswhite2018template}
Nate Crosswhite, Jeffrey Byrne, Chris Stauffer, Omkar Parkhi, Qiong Cao, and
  Andrew Zisserman.
\newblock Template adaptation for face verification and identification.
\newblock {\em Image and Vision Computing}, 79:35--48, 2018.

\bibitem{HOG}
N. Dalal and B. Triggs.
\newblock Histograms of oriented gradients for human detection.
\newblock In {\em 2005 IEEE Computer Society Conference on Computer Vision and
  Pattern Recognition (CVPR'05)}, volume~1, pages 886--893 vol. 1, 2005.

\bibitem{Arcface}
Jiankang Deng, Jia Guo, Niannan Xue, and Stefanos Zafeiriou.
\newblock Arcface: Additive angular margin loss for deep face recognition.
\newblock In {\em Proceedings of the IEEE/CVF Conference on Computer Vision and
  Pattern Recognition (CVPR)}, June 2019.

\bibitem{dosovitskiy2020image}
Alexey Dosovitskiy, Lucas Beyer, Alexander Kolesnikov, Dirk Weissenborn,
  Xiaohua Zhai, Thomas Unterthiner, Mostafa Dehghani, Matthias Minderer, Georg
  Heigold, Sylvain Gelly, et~al.
\newblock An image is worth 16x16 words: Transformers for image recognition at
  scale.
\newblock {\em arXiv preprint arXiv:2010.11929}, 2020.

\bibitem{Curricularface}
Yuge Huang, Yuhan Wang, Ying Tai, Xiaoming Liu, Pengcheng Shen, Shaoxin Li,
  Jilin Li, and Feiyue Huang.
\newblock Curricularface: Adaptive curriculum learning loss for deep face
  recognition.
\newblock In {\em Proceedings of the IEEE/CVF Conference on Computer Vision and
  Pattern Recognition (CVPR)}, June 2020.

\bibitem{multilossfusion}
Bhavin Jawade, Akshay Agarwal, Srirangaraj Setlur, and Nalini Ratha.
\newblock Multi loss fusion for matching smartphone captured contactless finger
  images.
\newblock In {\em 2021 IEEE International Workshop on Information Forensics and
  Security (WIFS)}, pages 1--6, 2021.

\bibitem{Ridegbase}
Bhavin Jawade, Deen~Dayal Mohan, Srirangaraj Setlur, Nalini Ratha, and Venu
  Govindaraju.
\newblock Ridgebase: A cross-sensor multi-finger contactless fingerprint
  dataset.
\newblock In {\em 2022 IEEE International Joint Conference on Biometrics
  (IJCB)}, pages 1--9, 2022.

\bibitem{videobasedreid}
Xinyang Jiang, Yifei Gong, Xiaowei Guo, Qize Yang, Feiyue Huang, Wei-Shi Zheng,
  Feng Zheng, and Xing Sun.
\newblock Rethinking temporal fusion for video-based person re-identification
  on semantic and time aspect.
\newblock In {\em Proceedings of the AAAI Conference on Artificial
  Intelligence}, volume~34, pages 11133--11140, 2020.

\bibitem{droneSURF}
Isha Kalra, Maneet Singh, Shruti Nagpal, Richa Singh, Mayank Vatsa, and P.~B.
  Sujit.
\newblock Dronesurf: Benchmark dataset for drone-based face recognition.
\newblock In {\em 2019 14th IEEE International Conference on Automatic Face and
  Gesture Recognition (FG 2019)}, pages 1--7, 2019.

\bibitem{khosla2020supervised}
Prannay Khosla, Piotr Teterwak, Chen Wang, Aaron Sarna, Yonglong Tian, Phillip
  Isola, Aaron Maschinot, Ce Liu, and Dilip Krishnan.
\newblock Supervised contrastive learning.
\newblock {\em Advances in neural information processing systems},
  33:18661--18673, 2020.

\bibitem{Adaface}
Minchul Kim, Anil~K. Jain, and Xiaoming Liu.
\newblock Adaface: Quality adaptive margin for face recognition.
\newblock In {\em Proceedings of the IEEE/CVF Conference on Computer Vision and
  Pattern Recognition (CVPR)}, pages 18750--18759, June 2022.

\bibitem{CAFace}
Minchul Kim, Feng Liu, Anil~K Jain, and Xiaoming Liu.
\newblock Cluster and aggregate: Face recognition with large probe set.
\newblock {\em Advances in Neural Information Processing Systems},
  35:36054--36066, 2022.

\bibitem{GAP}
Min Lin, Qiang Chen, and Shuicheng Yan.
\newblock Network in network.
\newblock {\em arXiv preprint arXiv:1312.4400}, 2013.

\bibitem{Sphereface}
Weiyang Liu, Yandong Wen, Zhiding Yu, Ming Li, Bhiksha Raj, and Le Song.
\newblock Sphereface: Deep hypersphere embedding for face recognition.
\newblock In {\em Proceedings of the IEEE Conference on Computer Vision and
  Pattern Recognition (CVPR)}, July 2017.

\bibitem{RSA}
Xiaofeng Liu, Zhenhua Guo, Site Li, Lingsheng Kong, Ping Jia, Jane You, and BVK
  Kumar.
\newblock Permutation-invariant feature restructuring for correlation-aware
  image set-based recognition.
\newblock In {\em Proceedings of the IEEE/CVF International Conference on
  Computer Vision}, pages 4986--4996, 2019.

\bibitem{MAGFace}
Qiang Meng, Shichao Zhao, Zhida Huang, and Feng Zhou.
\newblock Magface: A universal representation for face recognition and quality
  assessment.
\newblock In {\em Proceedings of the IEEE/CVF Conference on Computer Vision and
  Pattern Recognition}, pages 14225--14234, 2021.

\bibitem{LBP}
T. Ojala, M. Pietikainen, and T. Maenpaa.
\newblock Multiresolution gray-scale and rotation invariant texture
  classification with local binary patterns.
\newblock {\em IEEE Transactions on Pattern Analysis and Machine Intelligence},
  24(7):971--987, 2002.

\bibitem{VGGFace}
Omkar~M. Parkhi, Andrea Vedaldi, and Andrew Zisserman.
\newblock Deep face recognition.
\newblock In Xianghua Xie, Mark~W. Jones, and Gary K.~L. Tam, editors, {\em
  Proceedings of the British Machine Vision Conference (BMVC)}, pages
  41.1--41.12. BMVA Press, September 2015.

\bibitem{TADPool}
Nishant Sankaran, Deen~Dayal Mohan, Sergey Tulyakov, Srirangaraj Setlur, and
  Venugopal Govindaraju.
\newblock Tadpool: Target adaptive pooling for set based face recognition.
\newblock In {\em 2021 16th IEEE International Conference on Automatic Face and
  Gesture Recognition (FG 2021)}, pages 1--8, 2021.

\bibitem{MFAN}
Nishant Sankaran, Sergey Tulyakov, Srirangaraj Setlur, and Venu Govindaraju.
\newblock Metadata-based feature aggregation network for face recognition.
\newblock In {\em 2018 International Conference on Biometrics (ICB)}, pages
  118--123, 2018.

\bibitem{tulyakov2019utilizing}
Sergey Tulyakov, Nishant Sankaran, Srirangaraj Setlur, and Venu Govindaraju.
\newblock Utilizing template diversity for fusion of face recognizers.
\newblock In {\em 2019 IEEE 5th International Conference on Identity, Security,
  and Behavior Analysis (ISBA)}, pages 1--7. IEEE, 2019.

\bibitem{Cosface}
Hao Wang, Yitong Wang, Zheng Zhou, Xing Ji, Dihong Gong, Jingchao Zhou, Zhifeng
  Li, and Wei Liu.
\newblock Cosface: Large margin cosine loss for deep face recognition.
\newblock In {\em Proceedings of the IEEE Conference on Computer Vision and
  Pattern Recognition (CVPR)}, June 2018.

\bibitem{MV-Arc-Softmax}
Xiaobo Wang, Shifeng Zhang, Shuo Wang, Tianyu Fu, Hailin Shi, and Tao Mei.
\newblock Mis-classified vector guided softmax loss for face recognition.
\newblock {\em Proceedings of the AAAI Conference on Artificial Intelligence},
  34(07):12241--12248, Apr. 2020.

\bibitem{MCN}
Weidi Xie and Andrew Zisserman.
\newblock Multicolumn networks for face recognition.
\newblock {\em arXiv preprint arXiv:1807.09192}, 2018.

\bibitem{NAN}
Jiaolong Yang, Peiran Ren, Dongqing Zhang, Dong Chen, Fang Wen, Hongdong Li,
  and Gang Hua.
\newblock Neural aggregation network for video face recognition.
\newblock In {\em Proceedings of the IEEE conference on computer vision and
  pattern recognition}, pages 4362--4371, 2017.

\bibitem{MTCNN}
Kaipeng Zhang, Zhanpeng Zhang, Zhifeng Li, and Yu Qiao.
\newblock Joint face detection and alignment using multitask cascaded
  convolutional networks.
\newblock {\em IEEE signal processing letters}, 23(10):1499--1503, 2016.

\bibitem{Adacos}
Xiao Zhang, Rui Zhao, Yu Qiao, Xiaogang Wang, and Hongsheng Li.
\newblock Adacos: Adaptively scaling cosine logits for effectively learning
  deep face representations.
\newblock In {\em Proceedings of the IEEE/CVF Conference on Computer Vision and
  Pattern Recognition (CVPR)}, June 2019.

\bibitem{zhong2019ghostvlad}
Yujie Zhong, Relja Arandjelovi{\'c}, and Andrew Zisserman.
\newblock Ghostvlad for set-based face recognition.
\newblock In {\em Computer Vision--ACCV 2018: 14th Asian Conference on Computer
  Vision, Perth, Australia, December 2--6, 2018, Revised Selected Papers, Part
  II 14}, pages 35--50. Springer, 2019.

\bibitem{webface}
Zheng Zhu, Guan Huang, Jiankang Deng, Yun Ye, Junjie Huang, Xinze Chen, Jiagang
  Zhu, Tian Yang, Dalong Du, Jiwen Lu, et~al.
\newblock Webface260m: A benchmark for million-scale deep face recognition.
\newblock {\em IEEE Transactions on Pattern Analysis and Machine Intelligence},
  2022.

\end{thebibliography}
}

\end{document}